\documentclass[10pt,twocolumn,letterpaper]{article}

\usepackage{cvpr}
\usepackage{times}
\usepackage{epsfig}
\usepackage{graphicx}
\usepackage{amsmath}
\usepackage{amssymb}
\usepackage{algorithm}  
\usepackage{algorithmicx}  
\usepackage{algpseudocode}
\usepackage[normalem]{ulem}
\usepackage{ulem}
\usepackage{makecell,multirow,diagbox}
\usepackage{indentfirst}

\usepackage[pagebackref=true,breaklinks=true,letterpaper=true,colorlinks,bookmarks=false]{hyperref}

\cvprfinalcopy 

\def\cvprPaperID{3962} 

\begin{document}

\title{Unsupervised Part Mining for Fine-grained Image Classification}

\author{Runsheng Zhang\footnotemark[1],~Jian Zhang\footnotemark[1],~Yaping Huang and Qi Zou\\
Beijing Jiaotong University, Beijing, China\\
{\tt\small \{ rszhang, jianzhang1,yphuang, qzou\}@bjtu.edu.cn
}
}
\maketitle
\renewcommand{\thefootnote}{\fnsymbol{footnote}}
\footnotetext[1]{These authors contributed equally to this work and should be considered co-first authors}
\begin{abstract}
Fine-grained image classification remains challenging due to the large intra-class variance and small inter-class variance. Since the subtle visual differences are only in local regions of discriminative parts among subcategories, part localization is a key issue for fine-grained image classification. Most existing approaches localize object or parts in an image with object or part annotations, which are expensive and labor-consuming. To tackle this issue, we propose a fully unsupervised part mining (UPM) approach to localize the discriminative parts without even image-level annotations, which largely improves the fine-grained classification performance. We first utilize pattern mining techniques to discover frequent patterns, \emph{\ie}, co-occurrence highlighted regions, in the feature maps extracted from a pre-trained convolutional neural network (CNN) model. Inspired by the fact that these relevant meaningful patterns typically hold appearance and spatial consistency, we then cluster the mined regions to obtain the cluster centers and the discriminative parts surrounding the cluster centers are generated. Importantly, any annotations and sophisticated training procedures are not used in our proposed part localization approach. Finally, a multi-stream classification network is built for aggregating the original, object-level and part-level features simultaneously. Compared with other state-of-the-art approaches, our UPM approach achieves the competitive performance.
\end{abstract}

\section{Introduction}
Fine-grained image classification aims to recognize hundreds of subcategories belonging to a basic-level category (\eg, birds~\cite{wah2011caltech}, dogs~\cite{khosla2011novel}, cars~\cite{krause20133d} and aircrafts~\cite{maji2013fine}), which has attracted increasing attention in computer vision and pattern recognition.
Compared with general object classification, this task is extremely challenging due to the large variance in the same subcategory and small variance among different subcategories. 
Since these subcategories are similar in global appearances, different subcategories can only be distinguished by subtle visual differences existed in local regions of key parts, such as the shape of beak, the color of foot and the texture of feather for bird. Thus, localizing object and discriminative parts is highly essential for fine-grained image classification.

\begin{figure*}[t]
\begin{center}
\includegraphics[width=1\linewidth]{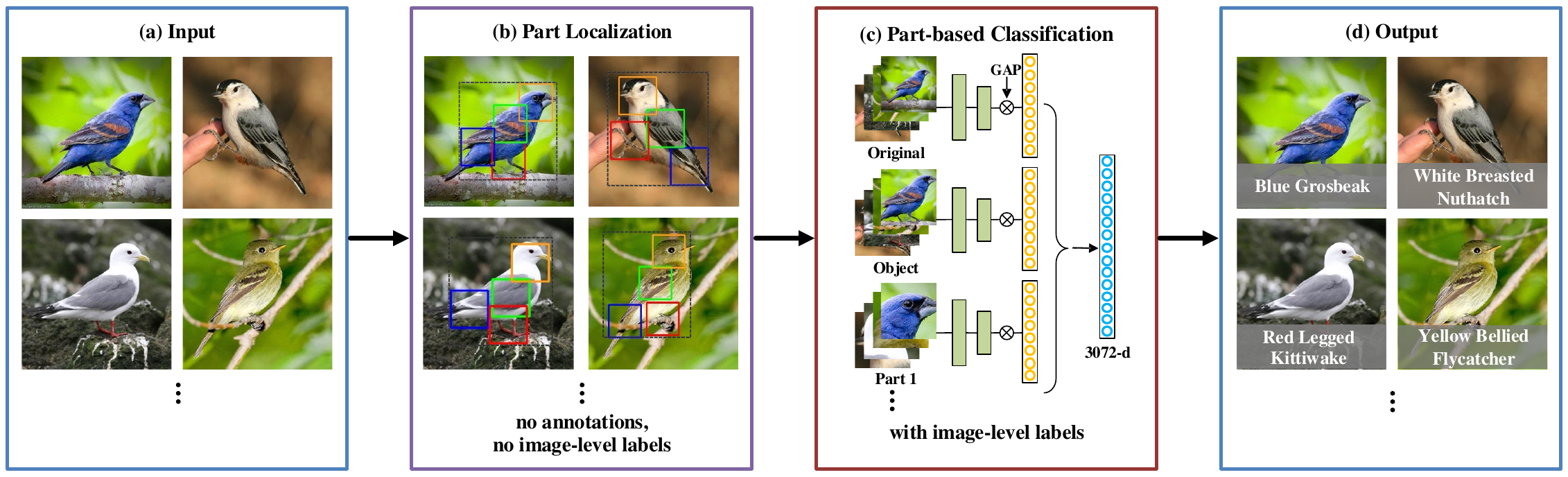}
\end{center}
   \caption{The overview of our part mining-based classification framework. Given an image, the discriminative parts are localized by our unsupervised part mining (UPM) approach in the purple box. Part-based classification network is a multi-stream network to aggregate different level features for final classification in the red box. Note that our UPM approach does not require any annotations even image-level labels in part localization module. (Best viewed in color)}
\label{fig:overview}
\end{figure*}

Inspiringly, a majority of fine-grained image classification methods have incorporated part localization and achieve significant progress. However, most earlier works~\cite{branson2014bird,huang2016part,wei2018mask,zhang2016spda,zhang2014part} still utilize strong supervision of human-labeled object annotation (\ie, bounding box of object) or part annotations (\ie, part locations). Since the object and part annotations are laborious and expensive, many works~\cite{xiao2015application,zhang2016weakly,zhang2016picking,liu2016fully,zhao2017diversified,fu2017look,zheng2017learning} address part localization under a weakly-supervised setting with only image-level labels.
Those methods can be roughly divided into two dimensions: two-stage methods which perform part localization and fine-grained classification separately, and end-to-end training methods which jointly learn discriminative part localization and fine-grained feature representation. Most of two-stage methods ~\cite{zhang2016weakly,xiao2015application,zhang2016picking,zhang2016detecting} use region proposals~\cite{uijlings2013selective} as candidate regions to localize the discriminative parts, which may lead to low accuracy and high time consumption. Recently, ~\cite{fu2017look, zheng2017learning} propose the end-to-end framework where part localization and feature learning could mutually reinforce each other. Although promising results have been reported, it is highly difficult to train the models due to sophisticated alternative training procedures.

To deal with the above issues, we propose an unsupervised part mining (UPM) approach for fine-grained image classification. Our proposed part localization method is fully unsupervised, without any annotations even image-level labels. The key idea of our proposed UPM is to explore the distinctive parts from the pattern mining perspective. To realize the idea, we reuse the pre-trained CNN model which has powerful abilities of representation, and further employ pattern mining techniques to effectively mine frequently-occurring visual patterns from a large number of CNN activations. These mined patterns are highly corresponding to the possible parts, which could be exploited to boost the classification performance. Our proposed UPM approach is simple but effective, which does not require complex and long-time training process. Meanwhile, we have no dependencies on any annotations including image-level labels, and thus it greatly increases the usability and scalability of fine-grained classification.


Our approach consists of a part localization module and a part-based classification module, as shown in Figure \ref{fig:overview}. In part localization module, we reuse a pre-trained CNN model and propose to employ pattern mining techniques for localizing the possible parts without using any annotations. Specifically, we first convert the deep features from multiple convolutional layers of a pre-trained CNN model (\eg, VGG-16~\cite{simonyan2014very}) into a set of transactions, and then discover the co-occurrence patterns through pattern mining techniques. We observe that the relevant patterns generally correspond to representative local regions in one image. Motivated by this observation, we utilize simple clustering algorithms (\eg, k-means) to cluster the mined patterns with frequency information into multiple clusters. Finally, the regions surrounding cluster centers are the key parts for a given image and can be further used for fine-grained image classification. In part-based classification module, these localized parts are further clustered based on deep features and fed into a deep classification network, in which a multi-stream architecture is built to aggregate different level features for subsequent fine-grained classification. Our main contributions can be summarized as follows:
\begin{itemize}
\item We present a novel and effective unsupervised part localization approach, without any image-level labels, which is the key issue for fine-grained image classification. The experimental results show that the localized parts contribute to the final classification accuracy.
\item To the best of our knowledge, we propose the first usage of pattern mining for fine-grained image classification successfully, which fully exploit information from convolutional activations of a pre-trained CNN model.
\item We conduct comprehensive experiments on three challenging fine-grained datasets (Caltech-UCSD Birds, Stanford Cars and FGVC-Aircraft), and achieve competitive performance compared with the state-of-the-art methods.
\end{itemize}

The rest of this paper is organized as follows. Section \ref{section2} briefly describes the related works. Section \ref{section3} introduces our proposed method, and Section \ref{section4} shows the evaluation as well as the analysis. Finally Section \ref{section5} concludes this paper.


\section{Related Work}\label{section2}
\subsection{Fine-grained Image Classification}
Fine-grained image classification is a fundamental and important task in computer vision, and a large amount of works have been developed in the past few years. Benefited from the advancement of deep learning, many works~\cite{krizhevsky2012imagenet,simonyan2014very,zhang2014part,xiao2015application,lin2015bilinear} learn more discriminative feature representation by leveraging deep CNNs, and achieve significant progress.


Since subtle visual differences mostly reside in local regions of parts, discriminative part localization is crucial for fine-grained image classification. There are numerous emerging works proceeding along part localization. \cite{zhang2014part,huang2016part,zhang2016spda,wei2018mask} learn accurate part localization models with manual object bounding boxes and part annotations. Considering that the annotations are laborious and expensive, some works~\cite{zhang2016weakly,zhang2016picking,he2017fine,xiao2015application,zhao2017diversified,liu2016fully,fu2017look,zheng2017learning} begin to focus on how to exploit parts under a weakly-supervised setting with only image-level labels. \cite{zhang2016picking} proposes an automatic fine-grained classification method, incorporating deep convolutional filters with significant and consistent responses for both parts selection and representation.
Some of the above part localization-based methods~\cite{zhang2016weakly,xiao2015application,zhang2016picking,zhang2016detecting} usually require to firstly produce object or part candidates by selective search~\cite{uijlings2013selective}, which poses challenges to accurate part localization. 

Additionally, some weakly-supervised methods~\cite{sermanet2014attention,xiao2015application,zhao2017diversified,liu2016fully,fu2017look,zheng2017learning} use visual attention mechanism to automatically capture the informative regions. \cite{liu2016fully} employs a fully convolutional attention network to adaptively localize multiple parts simultaneously. Recent works~\cite{fu2017look, zheng2017learning} propose the end-to-end framework where part localization and feature learning could mutually reinforce each other. Although promising results have been reported, it is highly difficult to train the models due to sophisticated alternative training procedures.

Compared with previous efforts, our UPM approach can accurately localize the parts in a fully unsupervised way without even image-level labels, thus it does not need sophisticated training procedures. Moreover, it also does not rely on enormous region proposals. In addition, it is worth to note that NAC~\cite{simon2015neural} also considers the part localization in a fully-unsupervised manner without image-level annotations, which is similar to our work. However, our proposed method can directly localize multiple fine-grained parts instead of selecting useful ones from part proposals, and outperform NAC by a large margin.
\subsection{Pattern mining in Computer Vision}
Pattern mining is one of the most intensively investigated problems in data mining domain. Generally, a set of patterns is a combination of several elements, which capture the distinctive information. Inspired by this fact, more researchers rise to investigate the problem of employing pattern mining to address computer vision tasks, including image classification~\cite{fernando2013mining,li2017mining}, image collection summarization~\cite{rematas2015dataset} and object retrieval~\cite{fernando2013mining}.

A key issue of pattern mining is how to transform an image into transactions, which could retain the discriminative information as much as possible and also guarantee that those transactions should be suitable for pattern mining. Earlier works~\cite{quack2007efficient,agarwal2008multilevel} simply treat an individual visual word as an item in a transaction by adopting local bag-of-words as image representation. \cite{fernando2014mining} proposes a frequent local histograms method to represent an image with the histograms of patterns sets. Recently, \cite{fernando2013mining} is a pioneering work to illustrate how pattern mining techniques are combined with the CNN features. In~\cite{fernando2013mining}, a local patch is transformed into a transaction by treating each dimension index of a CNN activation from fully-connected layer as an item.
\begin{figure*}[t]
\begin{center}
\includegraphics[width=1\linewidth]{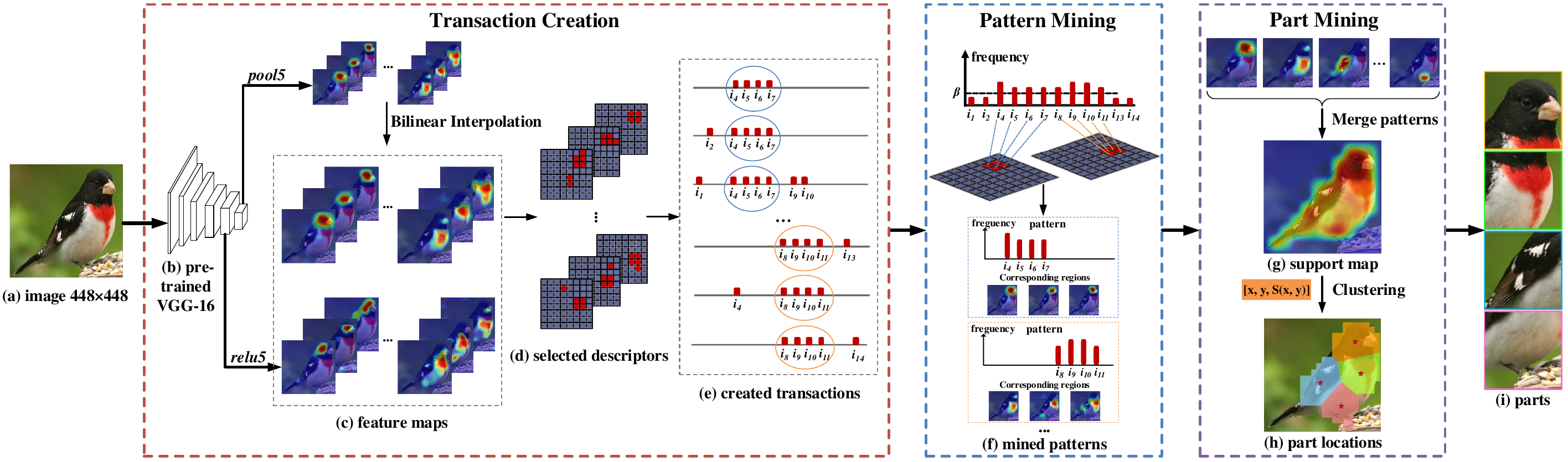}
\end{center}
   \caption{The pipeline of unsupervised part localization. First we feed an image in (a) into a pre-trained VGG-16~\cite{simonyan2014very} model in (b), and extract feature maps in (c) of $pool5$ and $relu5$ layers. We select the useful descriptors in (d) and convert them into items (\eg, $i_{1}$ in (e)). Each feature map is converted into a transaction in (e). $\{i_{1},i_{2}, ..., i_{14}\}$ is the index set of all highlighted positions, \ie, the item set of the transactions. Then we count the frequency of each item and remain the items whose frequency is greater than $\beta$. Thus, we mine the co-occurrence highlighted regions corresponding to the frequent patterns in (f). The mined patterns are merged to generate the support map in (g) and we perform clustering algorithm on it to obtain the part locations in (h). The outputs is the localized discriminative parts in (i). (Best viewed in color)}  
\label{fig:upmm}
\end{figure*}

\section{Approach}\label{section3}
In this section, we present the approach overview as shown in Figure \ref{fig:overview}. The approach is composed of an unsupervised part localization module (Section \ref{section3.2}) and a part-based classification module (Section \ref{section3.3}). In the first module, we aim to obtain the location of parts. The innovation of our approach is to localize discriminative parts by employing pattern mining techniques on the feature maps of a pre-trained CNN model. In the second module, we rely on the part locations to learn a joint feature representation and conduct part-based classification.

\subsection{Preliminary}
The following notations and terminology of data mining are used in the rest of this paper. Let $\textit{I}=\{i_{1}, i_{2},...,i_{\textit{M}}\}$ denotes an itemset containing \textit{M} items. A transaction \textit{T} is a subset of \textit{I} that satisfies to $|\textit{T}| \ll \textit{M}$, where $|\textit{T}|$ is the number of items in \textit{T}. A transaction database is defined as $\mathcal{D} = \{\textit{T}_{1}, \textit{T}_{2},...,\textit{T}_{\textit{N}}\}$, where $\forall i \in \{1,2,...,N\}, \textit{T}_{i} \in \mathcal{D}$. Given an itemset $\textit{P} \subseteq \textit{I}$, we define the support value of \textit{P} as:
\begin{equation}
supp(\textit{P}) = \frac{|\{\textit{T} \mid \textit{T} \in \mathcal{D}, \textit{P} \subseteq \textit{T}\}|}{N} \in [0,1],
\end{equation}
where $|\cdot|$ measures the cardinality. The support value of pattern \textit{P} indicates that how many transactions containing pattern \textit{P} in $\mathcal{D}$, \ie, $|\{\textit{T} \mid \textit{T} \in \mathcal{D}, \textit{P} \subseteq \textit{T}\}|$. \textit{P} is regarded as a frequent itemset when its support value is larger than a predefined threshold.

\subsection{Unsupervised Part Localization} \label{section3.2}
The goal of part localization is to obtain a collection of discriminative parts for a given fine-grained image. High-level convolutional layers can learn semantic cues, \ie, meaningful patterns, which correspond to whole objects~\cite{simonyan2013deep} or parts of objects~\cite{simon2014part}. Inspired by the observation, we propose a fully unsupervised part mining approach where the parts are discovered directly from activations of a pre-trained CNN model through pattern mining techniques without any labels. Note that the pre-trained model is not fine-tuned on the interest fine-grained dataset.

Figure \ref{fig:upmm} illustrates the pipeline of our UPM approach. We first extract feature maps from \textit{pool5} and \textit{relu5} layers of a pre-trained VGG-16~\cite{simonyan2014very} model, and then adopt pattern mining techniques to discover frequent patterns in these feature maps. Finally we perform the clustering algorithm on mined patterns and generate the parts surrounding the corresponding cluster centers. 


\subsubsection{Transaction Creation}
In order to apply pattern mining techniques to part localization task, the process of transforming the image into a set of transactions while retaining useful information is a key issue that must be tackled.

Given an input image \textbf{I}, we first feed it into a pre-trained VGG-16~\cite{simonyan2014very} model and extract feature maps from \textit{pool5} and \textit{relu5} layers in Figure \ref{fig:upmm} (c). We observe that most semantic parts of a bird are frequently fired at the same location in the feature maps. Moreover, the activations of two specific layers complement each other very well. Thus, we adopt a multi-layer combination strategy to alleviate the loss of useful information caused by only considering single layer activations. Besides, we need to resize $pool5$ feature maps to the same size of $relu5$ by bilinear interpolation, and we obtain 1,024 feature maps in total.


The dimension of each $relu5$ feature map is $h \times w$, where $h$ and $w$ indicate width and height of the feature map respectively. To simplify the process of creating transactions, we stretch each feature map into a vector $V \in \mathbb{R}^{h \times w}$.
In our UPM approach, each feature map is taken as a transaction \textit{$T$}, and each position index activated from the feature map is considered as an item $i_{j}$ ($j \in \{1,2,...,h \times w\}$). For example, if there are five positions activated from a feature map, the corresponding transaction contains five items denoted as $\textit{T}=\{i_{1}, i_{2}, i_{3}, i_{4}, i_{5}\}$. The set of all transactions is denoted as $\mathcal{D}$ and $\textit{T} \in \mathcal{D}$. The index set of all positions activated from feature maps, also known as an itemset, is denoted by $\textit{I}=\{i_{1}, i_{2},...,i_{\textit{m}}\}$. Generally, $\textit{T} \subseteq \textit{I}$.

Next, we select the meaningful descriptors in Figure \ref{fig:upmm} (d) to convert them into items. Specifically, we calculate the mean value $\alpha$ of the CNN activation responses larger than 0 as the tunable threshold instead of a fixed threshold in~\cite{li2017mining}. The position whose response value is higher than $\alpha$ is highlighted and its index will be converted into an item. Those indexes of all highlighted positions in one feature map finally form a transaction in Figure \ref{fig:upmm} (e).

\subsubsection{Pattern Mining}
Once a set of transactions $\mathcal{D}$ in Figure \ref{fig:upmm} (e) are created, we utilize the Apriori algorithm~\cite{agrawal1994fast} to discover frequent items (\ie, patterns). For a given minimum support threshold $\beta$, if $supp(\textit{P}) \ge \beta$, the itemset \textit{P} is considered as a pattern in Figure \ref{fig:upmm} (f). Note that the support value of the pattern indicates the frequency of this pattern appearing in all feature maps. Thus, the appropriate value of $\beta$ guarantees that we can mine the most representative and discriminative patterns.

\subsubsection{Part Mining}\label{section3.2.3}
Based on these mined patterns, we first select the largest connected component to remove those isolated patterns indicating background regions and merge the patterns to generate the support map. Subsequently, we  conduct clustering algorithm on the support map to localize multiple parts simultaneously. Finally, we adopt a simple and effective geometric constrains to crop a square surrounding each cluster center as a part region. Next, we present the details of part mining.

\textbf{Generating support map}. In our UPM approach, a mined pattern corresponds to a region in one image as shown in Figure \ref{fig:upmm} (f) and some relevant patterns generally indicate prominent representative local regions (\eg, the head of bird). Besides, we find that the isolated regions represented by one pattern or multiple patterns usually belong to the background of an image. Thus, we select the largest connected component based on all mined patterns to remove those isolated patterns. 

Here we introduce a new concept, support map, whose size is same with the feature map of \textit{relu5} layer. Note that the support map in Figure \ref{fig:upmm} (g) is generated by merging relevant and non-redundant patterns. Suppose that we have mined $n$ patterns denoted as $\{P_{1},P_{2},...,P_{n}\}$, the support map $S$ is defined as:
\begin{equation}
S(x,y) = \left\{
\begin{array}{lr}
f(x,y), & \text{if}~\exists P_{j}, i_{(x,y)} \in P_{j}, j \in [1,n] \\ 
0, & \text{otherwise}\\
\end{array}\label{supportmap}
\right.
\end{equation}
where $f(x,y)$ denotes the frequency of an item $i_{(x,y)}$ represented by its position $(x,y)$. To obtain the support map with the same size as the original image, we upsample the support map by bilinear interpolation. The support map indicates how many times each item $i_{(x,y)}$ would be activated from all feature maps. More importantly, the higher value $S(x,y)$ of the position, the more likely its corresponding region could be a part of the object.

\textbf{Finding part regions by clustering}. Inspired by the observation that some relevant patterns generally correspond to representative local regions (\eg, the head of bird) and the local regions are spatially continuous, thus we can divide the regions into several groups of spatial locations. An intuitive idea is to perform the clustering algorithm on the support map. Specifically, we first produce the clustering data, which are three-dimensional data including the coordinates of each spatial location $(x,y)$ and its corresponding support map value $S(x,y)$. Then we take them as input of the k-means algorithm to cluster these connected regions into $K$ clusters, as shown in Figure \ref{fig:upmm} (h). Surprisingly, the local regions represented by the patterns belonging to one cluster can be regarded as a discriminative part for a fine-grained image. Therefore, we obtain $K$ part locations $\textbf{C} = \{\textbf{c}_{1},\textbf{c}_{2},...,\textbf{c}_{i},...,\textbf{c}_{K}\}$ in the original image, where $\textbf{c}_{i}=(c_{ix},c_{iy})$ denotes the coordinates of the $i^{th}$ part.


After getting the part locations, then $K$ parts are generated by cropping $K$ squares from $\textbf{I}$, with each element of $\textbf{C}$ as the square center. However, if the side length of the part square is simply fixed, some cropped parts may only include a small part but be disturbed by large background noises. In addition, a fixed-size part may lead to serious overlap with other parts. Therefore, in order to tackle the issues and generate more representative and distinctive parts, we consider a simple and effective geometric constrains to determine the side length $l$ of a part as follows:
\begin{equation}
l = \lambda \times min\{w_{o},h_{o}\}
\label{eqn:part}
\end{equation}
where $w_{o}$ and $h_{o}$ are width and height of the bounding box generated from the support map respectively, and $\lambda$ is a scale factor. Finally, we can define the $i^{th}$ part region mask as:
\begin{equation}
\textbf{M}_{i}(x,y) = \left\{
\begin{array}{lr}
1, & |x - c_{ix}| \le \frac{l}{2},~|y - c_{iy}| \le \frac{l}{2}\\ 
0, & \text{otherwise}\\
\end{array}\label{eqn:part_mask}
\right.
\end{equation}
Thus, the $i^{th}$ cropped part region can be computed as:
\begin{equation}
\textbf{I}^{p_{i}}= \textbf{I} \odot \textbf{M}_{i}
\end{equation}
where $\odot$ denotes element-wise multiplication. Each part region is amplified into $224 \times 224$ for subsequent part-based classification. 

Algorithm \ref{algorithm:pattern clustering} gives the details of part mining.

\floatname{algorithm}{Algorithm}  
\renewcommand{\algorithmicrequire}{\textbf{Input:}}  
\renewcommand{\algorithmicensure}{\textbf{Output:}}
\begin{algorithm}
\caption{Part Mining}
\begin{algorithmic}[1] 
\Require \\
The input image $\textbf{I}$; \\
The set of mined patterns $\textbf{P} = \{P_{1},P_{2},...,P_{n}\}$; \\
The number of parts $K$.
\Ensure The masks of $K$ parts $\textbf{M}_{i}, i = \{1,2,...,K\}$.
\For {$i = 1,2,...,n$}
\State Map the items in pattern $P_{i}$ to a set of 2D coordinates X = $\{(x,y)\}$.
\For {each $(x,y) \in X$}
\State Compute support map value $S(x,y)$ according to Eqn. (\ref{supportmap}).
\EndFor
\EndFor
\State Use bilinear interpolation to upsample support map $S$ to the same size of $\textbf{I}$.
\State Label connected components for support map $S$.
\State Find a largest connected component $C$, and $S/C \leftarrow 0$.
\State Initialize $F \leftarrow \emptyset$.
\For {each spatial location $(x,y)$ of support map $S$}
\If {$S(x,y) > 0$}
\State $f=[x, y, S(x,y)]$.
\State $F \leftarrow F \cup {\{f\}}$.
\EndIf
\EndFor
\State Perform k-means algorithm on $F$ to obtain $K$ 2D part locations $\textbf{C} = \{\textbf{c}_{1}, \textbf{c}_{2}, ..., \textbf{c}_{K}\}$. 
\For {each $\textbf{c}_{i} \in \textbf{C}$}
\State Calculate the part mask $\textbf{M}_{i}$ according to Eqn. (\ref{eqn:part_mask}).
\EndFor \\
\Return $\textbf{M}_{i}, i = \{1, 2, ..., K\}$.
\end{algorithmic}\label{algorithm:pattern clustering}
\end{algorithm}
\subsection{Part-based Classification}\label{section3.3}
The different level focuses (\ie, image-level, object-level and part-level) have different representations and are complementary to improve the classification performance. Therefore, we build a multi-stream architecture with an \textit{Image stream}, an \textit{Object stream} and a \textit{Part stream} to learn a joint feature representation, as shown in Figure \ref{fig:overview}. Since previous works~\cite{zhang2014part,liu2016fully,fu2017look,zheng2017learning} indicate the benefits of region zooming, we amplify the original image to a higher resolution $448 \times 448$.  These images are taken as input to train a classification network based on the original image.

\textbf{Object stream}. Object localization can eliminate the influence of noisy background to learn representative object features. Thus, we also consider object localization in our method. Actually, we observe that the support map in Section \ref{section3.2} could indicate the representative object regions, as shown in Figure \ref{fig:upmm} (g). So it is reasonable to generate the object region from the support map. Specifically, we perform binarization and connectivity area extraction on the support map $S$, which is similar to CAM~\cite{zhou2016learning}. Finally, the images are cropped and resized into a fixed size of $448 \times 448$ to train a classification model based on the object-level images.

\textbf{Part stream}. Since the parts can capture the subtle and local discrimination within two similar subcategories, we train a set of classification models based on part-level images, each of which conducts classification on one part separately.

For the training set containing $N$ images, $N \times K$ parts are obtained by our UPM approach. However, these parts are out-of-order and not aligned by its semantic meaning. Therefore, we need to align these parts with the same semantic meaning together, so as to provide the training datasets for multiple part-level models. We are inspired by the fact that different convolutional layers learn different level features~\cite{zeiler2014visualizing}. Generally speaking, the higher deep convolutional layers carry more discriminative power and thus more likely to learn semantic cues (meaningful patterns, \eg bird's head or dog's face). 
An intuitive idea is that we can utilize clustering techniques to obtain the part clusters based on convolutional feature space. 

For clear expression, we denote the part mask as $\textbf{M}_{ij} (i = 1,2,...,K, j = 1,2,...,N)$, which represents the $i^{th}$ part mask of the $j^{th}$ training image. 
Specifically, we first extract convolutional features by feeding the original image \textbf{I} into a classification model trained on interest dataset (\eg, $\textit{conv5\_4}$ layer of VGG-19~\cite{simonyan2014very}). The extracted deep features are denoted as $\textbf{W} * \textbf{I}$, where $*$ represents a set of operations of convolution, pooling and activation, and $\textbf{W}$ represents the overall parameters of the model. Then we resize the part mask $\textbf{M}_{ij}$ in Section \ref{section3.2.3} to the same size of $\textbf{W} * \textbf{I}$. The features corresponding to the $i^{th}$ part region of the $j^{th}$ training image can be represented as:
\begin{equation}
\textbf{P}_{ij} = [\textbf{W} * \textbf{I}] \odot \textbf{M}_{ij}
\end{equation}

To reduce the dimension of features, global average pooling (GAP)~\cite{zhou2016learning} is performed on the above features. Finally, we obtain $N \times K$ feature descriptors and perform the spectral clustering algorithm on them to partition those corresponding parts into $K$ groups. Each part-level CNN model is fine-tuned on corresponding parts separately.


\textbf{Joint feature representation}: In our work, we leverage the feature ensemble strategy. The final feature representation can be represented as:
\begin{equation}
\{\textbf{P}_{or}, \textbf{P}_{ob}, \textbf{P}_{1}, \textbf{P}_{2}, ..., \textbf{P}_{K}\}
\end{equation}
where $\textbf{P}_{or}$, $\textbf{P}_{ob}$ and $\textbf{P}_{i}$ denote the feature descriptors of the original image, the object image and the $i^{th}$ part respectively. Each feature descriptor is extracted from the last convolutional layer of corresponding classification network. We first perform GAP and $\mathit{l}_{2}$-normalization on each feature descriptor, and concatenate them to train a classifier for the final classification.

\section{Experiment}\label{section4}
\subsection{Datasets}
To evaluate the effectiveness of our proposed method, we conduct experiments on three widely-used datasets for fine-grained image classification, including Caltech-UCSD Birds (CUB-200-2011)~\cite{wah2011caltech}, Stanford Cars~\cite{krause20133d} and FGVC-Aircraft~\cite{maji2013fine}. 

\subsection{Implementation Details}
In our unsupervised part localization module, the input image is resized to $448 \times 448$, and then fed into a publicly available VGG-16~\cite{simonyan2014very} model pre-trained on ImageNet to extract feature maps from \textit{relu5} and \textit{pool5} layers. The minimum support threshold $\beta$ is set to 0.07, 0.06 and 0.05 on CUB-200-2011, Stanford Cars and FGVC-Aircraft datasets respectively. The number of parts $K$ is set to 4. The $\lambda$ in Eqn. (\ref{eqn:part}) is empirically set to $\frac{1}{4}$, which makes the parts more representative.

In the part-based classification experiments, we use VGG-19~\cite{simonyan2014very} and ResNet-50~\cite{he2016deep} as the baseline models. We first train an image-level classification model based on full-size images of $448 \times 448$. Then, we adopt our proposed UPM approach to generate object-level and part-level training samples. Afterwards, we use these samples to fine-tune the image-level model to obtain an object-level model and four part-level models respectively. The input size of the object-level and part-level models are $448 \times 448$ and $224 \times 224$ respectively. The output of each CNN is extracted by GAP from the last convolutional layer 
to generate the $512\textrm{-}d$ feature descriptor in Section \ref{section3.3}. All feature descriptors are concatenated into a $3072\textrm{-}d$ representation to train a linear SVM classifier~\cite{fan2008liblinear} for classification. We run experiments with MatConvNet~\cite{vedaldi2015matconvnet} and Caffe~\cite{jia2014caffe}.

\begin{table}
\small
\begin{center}
\begin{tabular}{|c|c|c|c|c|}
\hline
\multirow{2}{*}{Method} & 
\multicolumn{3}{c|}{Anno. in part localization} &
\multirow{2}{*}{Acc.(\%)} \\
\cline{2-4} & Object & Part & Image & \\
\hline\hline
Part-RCNN~\cite{zhang2014part} & $\checkmark$ & $\checkmark$ & $\checkmark$ & 76.4  \\
PS-CNN~\cite{huang2016part} & $\checkmark$ & $\checkmark$ & $\checkmark$ & 76.6 \\
PA-CNN~\cite{krause2015fine} & $\checkmark$ & & & 82.8 \\
FCAN~\cite{liu2016fully} & $\checkmark$ & & $\checkmark$ & 84.7 \\
B-CNN~\cite{lin2015bilinear} & $\checkmark$ & & & 85.1 \\
SPDA-CNN~\cite{zhang2016spda} & $\checkmark$ & $\checkmark$ & $\checkmark$ & 85.1 \\
PN-CNN~\cite{branson2014bird} & $\checkmark$ & $\checkmark$ & & 85.4 \\
Mask-CNN~\cite{wei2018mask} & & $\checkmark$ & & 85.4 \\
TLAN~\cite{xiao2015application} & & & $\checkmark$ & 77.9 \\
DVAN~\cite{zhao2017diversified} & & & $\checkmark$ & 79.0 \\
FCAN~\cite{liu2016fully} & & & $\checkmark$ & 84.3 \\
PDFR~\cite{zhang2016picking} & & & $\checkmark$ & 84.5 \\
RA-CNN~\cite{fu2017look} & & & $\checkmark$ & 85.3 \\
OPAM~\cite{peng2018object} & & & $\checkmark$ & 85.8 \\
MA-CNN~\cite{zheng2017learning} & & & $\checkmark$ & 86.5 \\
MAMC~\cite{sun2018multi} & & & $\checkmark$ & 86.5 \\
VGG-19~\cite{simonyan2014very} & & & & 78.9 \\
NAC~\cite{simon2015neural} & & & & 81.0 \\
ResNet-50~\cite{he2016deep} & & & & 82.9 \\
ST-CNN~\cite{jaderberg2015spatial} & & & & 84.1 \\
B-CNN~\cite{lin2015bilinear} &  & & & 84.1 \\
\hline
UPM (VGG-19) & & & & \textbf{81.9} \\
UPM (ResNet-50) & & & & \textbf{85.4} \\
\hline
\end{tabular}
\end{center}
\caption{Comparison results on CUB-200-2011 dataset. ``Object'', ``Part'' and ``Image'' represent whether the method uses the bounding box annotations, part annotations and image-level labels in part localization.}
\label{table:cub_result}
\end{table}

\subsection{Experiment on CUB-200-2011}

\begin{figure*}[t]
\begin{center}
\includegraphics[width=0.95\linewidth]{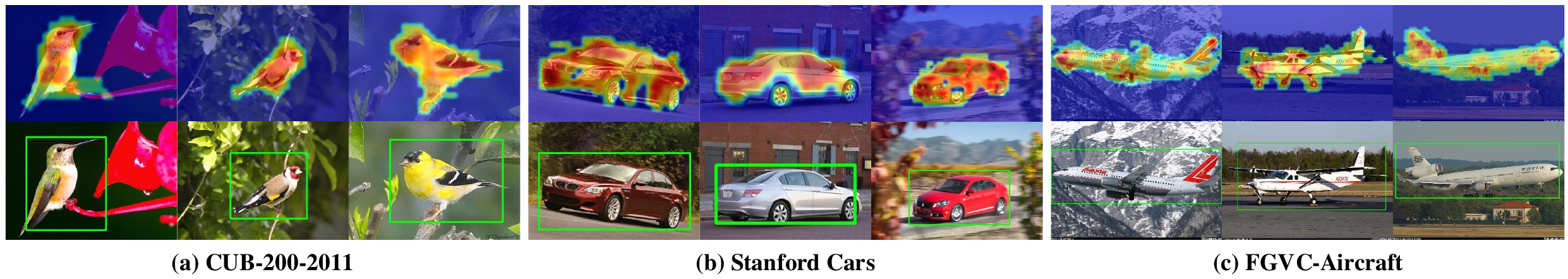}
\end{center}
   \caption{Examples of support maps and object localization results from (a) CUB-200-2011, (b) Standford Cars and (c) FGVC-Aircraft. The first row is the support maps, which indicate the representative object regions in images. The second row is the corresponding object localization results. (Best viewed in color)}  
\label{fig:object_localization}
\end{figure*}

\begin{figure*}[t]
\begin{center}
\includegraphics[width=0.95\linewidth]{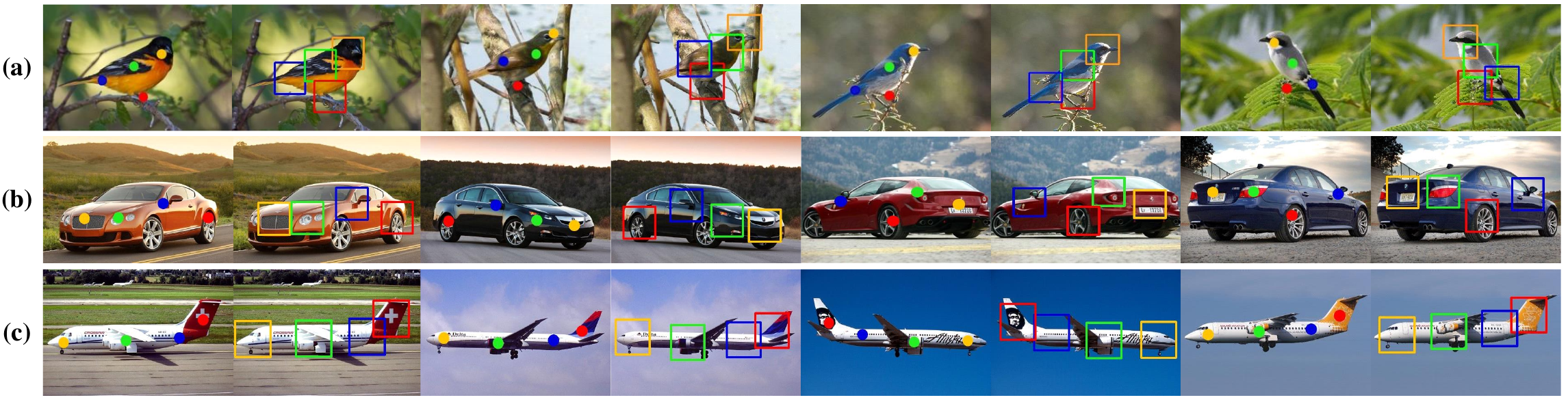}
\end{center}
   \caption{Examples of the part locations and corresponding part localization results from (a) CUB-200-2011, (b) Standford Cars and (c) FGVC-Aircraft. The four localized parts are discriminative to improve the classification results. Taking the bird in the first row as example, the yellow, green, blue and red dots dedicate the head, wing, tail and foot of the bird, respectively. (Best viewed in color)}
\label{fig:part_localization}
\end{figure*}

In this section, we compare our proposed UPM with the baseline methods and the state-of-the-arts on CUB-200-2011. The comparison results are summarized in Table \ref{table:cub_result}. 

Benefited from the localized parts by our UPM approach as shown in Figure \ref{fig:part_localization} (a), UPM (VGG-19) and UPM (ResNet-50) surpass the baseline models ResNet-50~\cite{he2016deep} and VGG-19~\cite{simonyan2014very} with 3.0\% and 2.5\% relative improvement respectively due to the effectiveness of part mining. Our approach outperforms most of the methods with strong supervision including bounding box, part annotation and image-level label listed in the Table \ref{table:cub_result}. Compared with the strong-supervised methods~\cite{zhang2016spda,branson2014bird,wei2018mask}, our approach achieves the comparable results without any annotations.

Compared with the weakly-supervised methods only with the image-level label, our approach is simple and does not need any annotations, but we still achieve comparable results. We outperform PDFR~\cite{zhang2016picking}, DVAN~\cite{zhao2017diversified} and FCAN~\cite{liu2016fully} by 0.9\%, 6.4\% and 1.1\% respectively. We are only lower 1.1\% than the recent MA-CNN~\cite{zheng2017learning} which jointly learns part proposals and feature representation.
However, our UPM approach can localize the parts in a fully unsupervised way even without image-level annotations, thus, unlike MA-CNN, we do not need the sophisticated training process.

UPM (ResNet-50) achieves the state-of-the-art results among methods under the same setting that are fully-unsupervised without any annotations. Compared with NAC~\cite{simon2015neural}, UPM (ResNet-50) achieves accuracy with 4.4\% relative improvement, which demonstrates that incorporating pattern mining techniques can efficiently mine the discriminative parts in an unsupervised manner.

\subsection{Experiment on Standford Cars}
We further evaluate the performance of our proposed method on the Standford Cars dataset. The results of part localization are shown in Figure \ref{fig:part_localization} (b).
The classification results are summarized in Table \ref{table:car_result}. 
UPM (ResNet-50) obtains 1.0\% higher accuracy than FCAN (with Object Anno.)~\cite{liu2016fully}. Besides, our approach achieves the competitive results compared with ~\cite{wang2016mining,krause2015fine}, which use bounding box annotations.
This benefits from the representativeness of support map and the effectiveness of pattern mining techniques.
Furthermore, our approach outperforms most of the weakly-supervised methods which use image-level labels, such as DVAN~\cite{zhao2017diversified}, FCAN (w/o Object Anno.)~\cite{liu2016fully} and OPAM~\cite{peng2018object}. Compared with FCAN (w/o Object Anno.)~\cite{liu2016fully}, the relative 3.2\% accuracy gain from UPM (ResNet-50) shows the significance of our mined parts in an unsupervised way.
Moreover, our approach surpasses B-CNN~\cite{lin2015bilinear}, which uses high dimensional features and requires image-level labels, with nearly 1.0\% relative accuracy gain.
\begin{table}
\small
\begin{center}
\begin{tabular}{|c|c|c|c|c|}
\hline
\multirow{2}{*}{Method} & 
\multicolumn{3}{c|}{Anno. in part localization} & 
\multirow{2}{*}{Acc.(\%)} \\
\cline{2-4} & Object & Part & Image & \\
\hline\hline
FCAN~\cite{liu2016fully} & $\checkmark$ & & $\checkmark$ & 91.3 \\
MDTP~\cite{wang2016mining} & $\checkmark$ & & $\checkmark$ & 92.5 \\
PA-CNN~\cite{krause2015fine} & $\checkmark$ & & & 92.8 \\
DVAN~\cite{zhao2017diversified} & & & $\checkmark$ & 87.1 \\
FCAN~\cite{liu2016fully} & & & $\checkmark$ & 89.1 \\
OPAM~\cite{peng2018object} & & & $\checkmark$ & 92.2 \\
RA-CNN~\cite{fu2017look} & & & $\checkmark$ & 92.5 \\
MA-CNN~\cite{zheng2017learning} & & & $\checkmark$ & 92.8 \\
MAMC~\cite{sun2018multi} & & & $\checkmark$ & 93.0 \\
VGG-19~\cite{simonyan2014very} & & & & 85.0 \\
ResNet-50~\cite{he2016deep} & & & & 89.6 \\
B-CNN~\cite{lin2015bilinear} &  & & & 91.3 \\
\hline
UPM (VGG-19) & & & & \textbf{89.2} \\
UPM (ResNet-50) & & & & \textbf{92.3} \\
\hline
\end{tabular}
\end{center}
\caption{Comparison results on Stanford Cars dataset. ``Object'', ``Part'' and ``Image'' represent whether the method uses the bounding box annotations, part annotations and image-level labels in part localization.}
\label{table:car_result}
\end{table}

\subsection{Experiment on FGVC-Aircraft}
Considering the simple background of aircraft images, we obtain good object localization results as shown in Figure \ref{fig:object_localization}. Therefore, the four localized parts are highly discriminative as shown in Figure \ref{fig:part_localization} (c). The classification results on FGVC-Aircraft dataset are summarized in Table \ref{table:air_result}. Our approach achieves superior performance over the state-of-the-art methods. Our approach outperforms our baseline models by 2.7\% and 3.3\%, respectively. Compared with MG-CNN~\cite{wang2015multiple} relying on object annotations, the 3.4\% clear margin from UPM (ResNet-50) shows the effectiveness of our UPM. 
We even surpass B-CNN (w/o Object Anno.)~\cite{lin2015bilinear} utilizing high dimensional features with nearly 5.9\% relative accuracy gains. 
It is worth to note that compared with MA-CNN~\cite{zheng2017learning} which relies on multiple alternative training stage, our approach can localize the parts in an unsupervised manner, but we still achieve better accuracy.

\begin{table}
\small
\begin{center}
\begin{tabular}{|c|c|c|c|c|}
\hline
\multirow{2}{*}{Method} & 
\multicolumn{3}{c|}{Anno. in part localization} &
\multirow{2}{*}{Acc.(\%)} \\
\cline{2-4} & Object & Part & Image & \\
\hline\hline
MG-CNN~\cite{wang2015multiple} & $\checkmark$ & & $\checkmark$ & 86.6 \\
MDTP~\cite{wang2016mining} & $\checkmark$ & & $\checkmark$ & 88.4 \\
MG-CNN~\cite{wang2015multiple} & & & $\checkmark$ & 82.5 \\
MA-CNN~\cite{zheng2017learning} & & & $\checkmark$ & 89.9 \\
FV-CNN~\cite{gosselin2014revisiting} & & & & 81.5 \\
VGG-19~\cite{simonyan2014very} & & & & 83.2 \\
B-CNN~\cite{lin2015bilinear} &  & & & 84.1 \\
ResNet-50~\cite{he2016deep} & & & & 86.7 \\
\hline
UPM (VGG-19) & & & & \textbf{85.9} \\
UPM (ResNet-50) & & & & \textbf{90.0} \\
\hline
\end{tabular}
\end{center}
\caption{Comparison results on FGVC-Aircraft dataset. ``Object'', ``Part'' and ``Image'' represent whether the method uses the bounding box annotations, part annotations and image-level labels in part localization.}
\label{table:air_result}
\end{table}

\subsection{Further Analysis}
We further show the quantitative comparison in Table \ref{table:further} to verify the performance of the streams used in our UPM approach. 
We can observe that our UPM (ResNet-50) approach outperforms the ``Original-stream+Object-stream''
with 1.0\% relative gains due to the complementarity with the original and object image, which shows the effectiveness of the localized parts through our UPM approach.

\begin{table}
\small
\begin{center}
\begin{tabular}{|c|c|}
\hline
Method & Acc.(\%) \\
\hline\hline
\textbf{Our UPM (ResNet-50) approach} & \multirow{2}{*}{\textbf{85.4}} \\ 
\textbf{(Original-stream+Object-stream+Part-stream)} & \\ \hline
Original-stream & 82.9 \\
Original-stream+Object-stream & 84.4 \\
\hline
\end{tabular}
\end{center}
\caption{Performance of different streams in our UPM approach on CUB-200-2011.}
\label{table:further}
\end{table}

\section{Conclusions}\label{section5}
In this paper, we propose a fully unsupervised part mining approach for fine-grained image classification, which explores the discriminative parts by incorporating the pattern mining techniques. We employ the pattern mining techniques to discover frequent patterns in the feature maps extracted from a pre-trained CNN model and perform the clustering algorithm on mined patterns to generate the parts. The proposed approach does not require any annotations even image-level labels in part localization, and does not require sophisticated training procedures. 
Extensive experiments show the effectiveness of UPM compared with other state-of-the-arts on three challenging fine-grained datasets.

{\small
\bibliographystyle{ieee}
\bibliography{egbib}

\begin{thebibliography}{10}\itemsep=-1pt

\bibitem{agarwal2008multilevel}
A.~Agarwal and B.~Triggs.
\newblock Multilevel image coding with hyperfeatures.
\newblock {\em International Journal of Computer Vision}, 78(1):15--27, 2008.

\bibitem{agrawal1994fast}
R.~Agrawal, R.~Srikant, et~al.
\newblock Fast algorithms for mining association rules.
\newblock In {\em Proc. 20th int. conf. very large data bases, VLDB}, volume
  1215, pages 487--499, 1994.

\bibitem{branson2014bird}
S.~Branson, G.~Van~Horn, S.~Belongie, and P.~Perona.
\newblock Bird species categorization using pose normalized deep convolutional
  nets.
\newblock {\em arXiv preprint arXiv:1406.2952}, 2014.

\bibitem{fan2008liblinear}
R.-E. Fan, K.-W. Chang, C.-J. Hsieh, X.-R. Wang, and C.-J. Lin.
\newblock Liblinear: A library for large linear classification.
\newblock {\em Journal of machine learning research}, 9(Aug):1871--1874, 2008.

\bibitem{fernando2014mining}
B.~Fernando, E.~Fromont, and T.~Tuytelaars.
\newblock Mining mid-level features for image classification.
\newblock {\em International Journal of Computer Vision}, 108(3):186--203,
  2014.

\bibitem{fernando2013mining}
B.~Fernando and T.~Tuytelaars.
\newblock Mining multiple queries for image retrieval: On-the-fly learning of
  an object-specific mid-level representation.
\newblock In {\em Proceedings of the IEEE International Conference on Computer
  Vision}, pages 2544--2551, 2013.

\bibitem{fu2017look}
J.~Fu, H.~Zheng, and T.~Mei.
\newblock Look closer to see better: Recurrent attention convolutional neural
  network for fine-grained image recognition.
\newblock In {\em CVPR}, volume~2, page~3, 2017.

\bibitem{gosselin2014revisiting}
P.-H. Gosselin, N.~Murray, H.~J{\'e}gou, and F.~Perronnin.
\newblock Revisiting the fisher vector for fine-grained classification.
\newblock {\em Pattern Recognition Letters}, 49:92--98, 2014.

\bibitem{he2016deep}
K.~He, X.~Zhang, S.~Ren, and J.~Sun.
\newblock Deep residual learning for image recognition.
\newblock In {\em Proceedings of the IEEE conference on computer vision and
  pattern recognition}, pages 770--778, 2016.

\bibitem{he2017fine}
X.~He, Y.~Peng, and J.~Zhao.
\newblock Fine-grained discriminative localization via saliency-guided faster
  r-cnn.
\newblock In {\em Proceedings of the 2017 ACM on Multimedia Conference}, pages
  627--635. ACM, 2017.

\bibitem{huang2016part}
S.~Huang, Z.~Xu, D.~Tao, and Y.~Zhang.
\newblock Part-stacked cnn for fine-grained visual categorization.
\newblock In {\em Proceedings of the IEEE Conference on Computer Vision and
  Pattern Recognition}, pages 1173--1182, 2016.

\bibitem{jaderberg2015spatial}
M.~Jaderberg, K.~Simonyan, A.~Zisserman, et~al.
\newblock Spatial transformer networks.
\newblock In {\em Advances in neural information processing systems}, pages
  2017--2025, 2015.

\bibitem{jia2014caffe}
Y.~Jia, E.~Shelhamer, J.~Donahue, S.~Karayev, J.~Long, R.~Girshick,
  S.~Guadarrama, and T.~Darrell.
\newblock Caffe: Convolutional architecture for fast feature embedding.
\newblock In {\em Proceedings of the 22nd ACM international conference on
  Multimedia}, pages 675--678. ACM, 2014.

\bibitem{khosla2011novel}
A.~Khosla, N.~Jayadevaprakash, B.~Yao, and F.-F. Li.
\newblock Novel dataset for fine-grained image categorization: Stanford dogs.
\newblock In {\em Proc. CVPR Workshop on Fine-Grained Visual Categorization
  (FGVC)}, volume~2, page~1, 2011.

\bibitem{krause2015fine}
J.~Krause, H.~Jin, J.~Yang, and L.~Fei-Fei.
\newblock Fine-grained recognition without part annotations.
\newblock In {\em Proceedings of the IEEE Conference on Computer Vision and
  Pattern Recognition}, pages 5546--5555, 2015.

\bibitem{krause20133d}
J.~Krause, M.~Stark, J.~Deng, and L.~Fei-Fei.
\newblock 3d object representations for fine-grained categorization.
\newblock In {\em Proceedings of the IEEE International Conference on Computer
  Vision Workshops}, pages 554--561, 2013.

\bibitem{krizhevsky2012imagenet}
A.~Krizhevsky, I.~Sutskever, and G.~E. Hinton.
\newblock Imagenet classification with deep convolutional neural networks.
\newblock In {\em Advances in neural information processing systems}, pages
  1097--1105, 2012.

\bibitem{li2017mining}
Y.~Li, L.~Liu, C.~Shen, and A.~Van Den~Hengel.
\newblock Mining mid-level visual patterns with deep cnn activations.
\newblock {\em International Journal of Computer Vision}, 121(3):344--364,
  2017.

\bibitem{lin2015bilinear}
T.-Y. Lin, A.~RoyChowdhury, and S.~Maji.
\newblock Bilinear cnn models for fine-grained visual recognition.
\newblock In {\em Proceedings of the IEEE International Conference on Computer
  Vision}, pages 1449--1457, 2015.

\bibitem{liu2016fully}
X.~Liu, T.~Xia, J.~Wang, Y.~Yang, F.~Zhou, and Y.~Lin.
\newblock Fully convolutional attention networks for fine-grained recognition.
\newblock {\em arXiv preprint arXiv:1603.06765}, 2016.

\bibitem{maji2013fine}
S.~Maji, E.~Rahtu, J.~Kannala, M.~Blaschko, and A.~Vedaldi.
\newblock Fine-grained visual classification of aircraft.
\newblock {\em arXiv preprint arXiv:1306.5151}, 2013.

\bibitem{peng2018object}
Y.~Peng, X.~He, and J.~Zhao.
\newblock Object-part attention model for fine-grained image classification.
\newblock {\em IEEE Transactions on Image Processing}, 27(3):1487--1500, 2018.

\bibitem{quack2007efficient}
T.~Quack, V.~Ferrari, B.~Leibe, and L.~Van~Gool.
\newblock Efficient mining of frequent and distinctive feature configurations.
\newblock In {\em Computer Vision, 2007. ICCV 2007. IEEE 11th International
  Conference on}, pages 1--8. IEEE, 2007.

\bibitem{rematas2015dataset}
K.~Rematas, B.~Fernando, F.~Dellaert, and T.~Tuytelaars.
\newblock Dataset fingerprints: Exploring image collections through data
  mining.
\newblock In {\em Proceedings of the IEEE Conference on Computer Vision and
  Pattern Recognition}, pages 4867--4875, 2015.

\bibitem{sermanet2014attention}
P.~Sermanet, A.~Frome, and E.~Real.
\newblock Attention for fine-grained categorization.
\newblock {\em arXiv preprint arXiv:1412.7054}, 2014.

\bibitem{simon2015neural}
M.~Simon and E.~Rodner.
\newblock Neural activation constellations: Unsupervised part model discovery
  with convolutional networks.
\newblock In {\em Proceedings of the IEEE International Conference on Computer
  Vision}, pages 1143--1151, 2015.

\bibitem{simon2014part}
M.~Simon, E.~Rodner, and J.~Denzler.
\newblock Part detector discovery in deep convolutional neural networks.
\newblock In {\em Asian Conference on Computer Vision}, pages 162--177.
  Springer, 2014.

\bibitem{simonyan2013deep}
K.~Simonyan, A.~Vedaldi, and A.~Zisserman.
\newblock Deep inside convolutional networks: Visualising image classification
  models and saliency maps.
\newblock {\em arXiv preprint arXiv:1312.6034}, 2013.

\bibitem{simonyan2014very}
K.~Simonyan and A.~Zisserman.
\newblock Very deep convolutional networks for large-scale image recognition.
\newblock {\em arXiv preprint arXiv:1409.1556}, 2014.

\bibitem{sun2018multi}
M.~Sun, Y.~Yuan, F.~Zhou, and E.~Ding.
\newblock Multi-attention multi-class constraint for fine-grained image
  recognition.
\newblock {\em arXiv preprint arXiv:1806.05372}, 2018.

\bibitem{uijlings2013selective}
J.~R. Uijlings, K.~E. Van De~Sande, T.~Gevers, and A.~W. Smeulders.
\newblock Selective search for object recognition.
\newblock {\em International journal of computer vision}, 104(2):154--171,
  2013.

\bibitem{vedaldi2015matconvnet}
A.~Vedaldi and K.~Lenc.
\newblock Matconvnet: Convolutional neural networks for matlab.
\newblock In {\em Proceedings of the 23rd ACM international conference on
  Multimedia}, pages 689--692. ACM, 2015.

\bibitem{wah2011caltech}
C.~Wah, S.~Branson, P.~Welinder, P.~Perona, and S.~Belongie.
\newblock The caltech-ucsd birds-200-2011 dataset.
\newblock 2011.

\bibitem{wang2015multiple}
D.~Wang, Z.~Shen, J.~Shao, W.~Zhang, X.~Xue, and Z.~Zhang.
\newblock Multiple granularity descriptors for fine-grained categorization.
\newblock In {\em Proceedings of the IEEE International Conference on Computer
  Vision}, pages 2399--2406, 2015.

\bibitem{wang2016mining}
Y.~Wang, J.~Choi, V.~Morariu, and L.~S. Davis.
\newblock Mining discriminative triplets of patches for fine-grained
  classification.
\newblock In {\em Proceedings of the IEEE Conference on Computer Vision and
  Pattern Recognition}, pages 1163--1172, 2016.

\bibitem{wei2018mask}
X.-S. Wei, C.-W. Xie, J.~Wu, and C.~Shen.
\newblock Mask-cnn: Localizing parts and selecting descriptors for fine-grained
  bird species categorization.
\newblock {\em Pattern Recognition}, 76:704--714, 2018.

\bibitem{xiao2015application}
T.~Xiao, Y.~Xu, K.~Yang, J.~Zhang, Y.~Peng, and Z.~Zhang.
\newblock The application of two-level attention models in deep convolutional
  neural network for fine-grained image classification.
\newblock In {\em Proceedings of the IEEE Conference on Computer Vision and
  Pattern Recognition}, pages 842--850, 2015.

\bibitem{zeiler2014visualizing}
M.~D. Zeiler and R.~Fergus.
\newblock Visualizing and understanding convolutional networks.
\newblock In {\em European conference on computer vision}, pages 818--833.
  Springer, 2014.

\bibitem{zhang2016spda}
H.~Zhang, T.~Xu, M.~Elhoseiny, X.~Huang, S.~Zhang, A.~Elgammal, and D.~Metaxas.
\newblock Spda-cnn: Unifying semantic part detection and abstraction for
  fine-grained recognition.
\newblock In {\em Proceedings of the IEEE Conference on Computer Vision and
  Pattern Recognition}, pages 1143--1152, 2016.

\bibitem{zhang2016detecting}
L.~Zhang, Y.~Yang, M.~Wang, R.~Hong, L.~Nie, and X.~Li.
\newblock Detecting densely distributed graph patterns for fine-grained image
  categorization.
\newblock {\em IEEE Transactions on Image Processing}, 25(2):553--565, 2016.

\bibitem{zhang2014part}
N.~Zhang, J.~Donahue, R.~Girshick, and T.~Darrell.
\newblock Part-based r-cnns for fine-grained category detection.
\newblock In {\em European conference on computer vision}, pages 834--849.
  Springer, 2014.

\bibitem{zhang2016picking}
X.~Zhang, H.~Xiong, W.~Zhou, W.~Lin, and Q.~Tian.
\newblock Picking deep filter responses for fine-grained image recognition.
\newblock In {\em Proceedings of the IEEE Conference on Computer Vision and
  Pattern Recognition}, pages 1134--1142, 2016.

\bibitem{zhang2016weakly}
Y.~Zhang, X.-S. Wei, J.~Wu, J.~Cai, J.~Lu, V.-A. Nguyen, and M.~N. Do.
\newblock Weakly supervised fine-grained categorization with part-based image
  representation.
\newblock {\em IEEE Transactions on Image Processing}, 25(4):1713--1725, 2016.

\bibitem{zhao2017diversified}
B.~Zhao, X.~Wu, J.~Feng, Q.~Peng, and S.~Yan.
\newblock Diversified visual attention networks for fine-grained object
  classification.
\newblock {\em IEEE Transactions on Multimedia}, 19(6):1245--1256, 2017.

\bibitem{zheng2017learning}
H.~Zheng, J.~Fu, T.~Mei, and J.~Luo.
\newblock Learning multi-attention convolutional neural network for
  fine-grained image recognition.
\newblock In {\em Int. Conf. on Computer Vision}, volume~6, 2017.

\bibitem{zhou2016learning}
B.~Zhou, A.~Khosla, A.~Lapedriza, A.~Oliva, and A.~Torralba.
\newblock Learning deep features for discriminative localization.
\newblock In {\em Proceedings of the IEEE Conference on Computer Vision and
  Pattern Recognition}, pages 2921--2929, 2016.

\end{thebibliography}
}

\end{document}